%% file: main.tex
\documentclass{article}
\usepackage{cite}
\usepackage{amsmath,amssymb,amsfonts}
\usepackage{algorithmic}
\usepackage[dvipdfmx]{graphicx}
\usepackage{textcomp}
\usepackage{xcolor}
\usepackage{subfigure}
\usepackage[colorlinks=true]{hyperref}
\hypersetup{breaklinks=true}
\usepackage[hyphens]{xurl}
\usepackage{listings}
\usepackage[version=3]{mhchem}
\usepackage{multirow}
\usepackage{multicol}
\usepackage{pifont}
\usepackage{tikz}
\usepackage[normalem]{ulem}
\usepackage{rotating}
\input{macro.tex}
\input{performance.tex}

\begin{document}

\title{
  \lamm{}: Semi-Supervised Pre-Training of\\Large-Scale Materials Models
}

\author{
  Yosuke Oyama, Yusuke Majima, Eiji Ohta, Yasufumi Sakai \\
  \textit{Fujitsu Limited} \\
  Kawasaki, Kanagawa \\
  211--8588, Japan \\
  \{oyama.yosuke, majima.yusuke, eiji, sakaiyasufumi\}@fujitsu.com
}

\maketitle

\begin{abstract}
  Neural network potentials (NNPs) are crucial for accelerating computational materials science by surrogating density functional theory (DFT) calculations.
  Improving their accuracy is possible through pre-training and fine-tuning, where an NNP model is first pre-trained on a large-scale dataset and then fine-tuned on a smaller, target dataset.
  However, this approach is computationally expensive, mainly due to the cost of DFT-based dataset labeling and load imbalances during large-scale pre-training.
  To address this, we propose LaMM, a semi-supervised pre-training method incorporating improved denoising self-supervised learning and a load-balancing algorithm for efficient multi-node training.
  We demonstrate that our approach effectively leverages a large-scale dataset of $\sim$300 million semi-labeled samples to train a single NNP model, resulting in improved fine-tuning performance in terms of both speed and accuracy.
\end{abstract}

\input{1_introduction.tex}
\input{2_related.tex}
\input{3_proposal.tex}
\input{4_evaluation.tex}
\input{5_conclusion.tex}

\section*{Acknowledgement}
This research used computational resources of AI Bridging Cloud Infrastructure (ABCI) provided by National Institute of Advanced Industrial Science and Technology (AIST).

\bibliography{reference}
\bibliographystyle{unsrt}

\end{document}

%% file: macro.tex
\newcommand{\lamm}[0]{LaMM}
\newcommand{\lamms}[0]{LaMM-S}
\newcommand{\lamml}[0]{LaMM-L}
\newcommand{\painn}[0]{PaiNN}
\newcommand{\eqvt}[0]{EquiformerV2}
\newcommand{\fairchem}[0]{FAIR-Chem}

\newcommand{\figref}[1]{Figure \ref{#1}}
\newcommand{\tabref}[1]{Table \ref{#1}}
\newcommand{\secref}[1]{Section \ref{#1}}

\renewcommand{\checkmark}{\ding{51}}
\newcommand{\crossmark}{\ding{55}}

\newcommand{\bestperf}[1]{\textbf{#1}}

%% file: performance.tex
\newcommand{\loadbalancinglammsspeedup}[0]{1.80x}

\newcommand{\loadbalancinglammsthroughput}[0]{2.44x}

\newcommand{\loadbalancinglammlthroughput}[0]{3.38x}

%% file: 1_introduction.tex
\section{Introduction} \label{section:introduction}
Materials informatics (MI) is a collective term for applying machine learning and data science approaches to material science.
Recently, as a part of MI, deep learning-based approaches have emerged as surrogate models for density functional theory (DFT), one of the most fundamental methods in computational material science.
DFT is an ab initio method for computing the electronic state of a set of atoms.
While DFT has many applications, such as molecular dynamics, its computational expense remains a challenge despite its accuracy.
Therefore, there is a high demand to accelerate DFT by using deep learning.

Machine learning interatomic potentials, or neural network potentials (NNPs), are surrogate models for DFT~\cite{schutt2021equivariant, deng2023chgnet, liao2024equiformerv, chen2022universal, yang2024mattersimdeeplearningatomistic, batzner20223, TAKAMOTO2022111280, NIPS2017_303ed4c6, ying2021transformers}.
An NNP model takes atoms' coordinates and atomic numbers as input and predicts their energy and per-atom forces.
While some NNP models include heads to predict forces directly, forces can also be naturally predicted by applying auto-differentiation to the energy with respect to the corresponding coordinates.
Many models adopt graph neural networks, in which atom coordinates are converted to a graph whose nodes correspond to atoms, and each edge represents a flag indicating that the two atoms are within a given cutoff distance.
Several DFT-based labeled datasets have been proposed as training datasets for NNP models.

One of the challenges in creating an NNP is that it requires a training dataset computed by DFT, which incurs a significant computational cost.
For example, OC20 and OMat24, two existing large datasets containing more than 100M DFT data samples, took 200 and 400 million hours to compute, respectively~\cite{das2021open,barrosoluque2024openmaterials2024omat24}.
Several large-scale pre-training efforts of NNP models have been performed in the last few years to tackle this challenge.
Although datasets for such pre-training may only cover areas of interest to some materials experts, such pre-trained models can be fine-tuned to provide accurate predictions for specific materials with limited dataset generation cost (i.e., DFT).
Examples of pre-trained NNP models are explained in \secref{subsection:pretrained_models}.

The current limitations of existing NNP pre-training approaches are as follows:
\begin{itemize}
\item Existing approaches focus on labeled datasets, where labels are computed by DFT. Combining labeled and unlabeled datasets has yet to be investigated in detail.
  Utilizing unlabeled datasets, which only contain materials' structural information, increases the pre-training dataset size by an order of magnitude without incurring expensive DFT labeling costs.
  Also, existing pre-training is limited in that all data samples have total energy and per-atom force labels, excluding some unique datasets with other sets of labels.
\item Mixing heterogeneous materials with respect to their system size (the number of atoms) results in poor computational performance in pre-training because of load imbalance.
  This imbalance is critical, especially for large models such as \eqvt{}, whose GPU batch size is limited to 10 or less due to its memory footprint.
\end{itemize}

In this work, we propose \lamm{}, a novel pre-training approach for NNPs utilizing a total of $\sim$300M organic and inorganic data samples.
Our work results in two different pre-trained models with different throughput-accuracy tradeoffs, \lamms{} and \lamml{}, whose base models are \painn{} and \eqvt{}, respectively.
We demonstrate that the pre-trained models improve not only fine-tuning time but also final accuracy for various downstream datasets.

Contributions of this work are as follows:

\begin{itemize}
\item We propose a semi-supervised learning methodology to utilize both DFT-labeled and unlabeled datasets.
  This methodology enables NNP pre-training with any combination of heterogeneous datasets, accepting any type of per-system and per-atom labels, or no labels at all.
  We combine various open datasets into a 300M-sample unified dataset by using this method.
\item We introduce a simple technique to improve the quality of semi-supervised coordinate denoising tasks for NNPs.
\item We propose a load-balancing technique for multi-GPU pre-training for NNPs, achieving \loadbalancinglammsthroughput{} and \loadbalancinglammlthroughput{} throughput improvement on 16 NVIDIA V100 GPUs for \painn{} and \eqvt{}, respectively.
\item We demonstrate that our pre-trained models accelerate fine-tuning and also improve inference accuracy on the HME21 benchmark dataset~\cite{takamoto2022towards}.
\end{itemize}

%% file: 2_related.tex
\section{Related work}

As briefly explained in \secref{section:introduction}, NNP models, typically GNNs, use atoms' coordinates and atomic numbers to predict per-system energy and forces.
This section explains the typical architecture of NNPs and prior work on large-scale training for generalized pre-trained models.

\subsection{NNP architecture}
A typical NNP model consists of two components: an encoder $f_\textrm{encoder}$ (or a backbone), typically a chain of interaction blocks, which computes atomic or edge (relationship between atoms) representations, and a set of heads (decoders) $f_\textrm{head\_*}$ that predict properties such as energy and forces \cite{schutt2021equivariant, deng2023chgnet, liao2024equiformerv, chen2022universal, yang2024mattersimdeeplearningatomistic, batzner20223, TAKAMOTO2022111280, NIPS2017_303ed4c6, ying2021transformers}:

\begin{eqnarray}
  s_i &=& f_\textrm{encoder}(\{\vec{r}_j\}_{j=1}^{N}, \{Z_j\}_{j=1}^{N}, i) \\
  E_i &=& f_\textrm{head\_energy}(s_i) \\
  E &=& \sum_{i=1}^{N} E_i \\
  \vec{F}_i &=& f_\textrm{head\_force}(s_i),
\end{eqnarray}
where $\vec{r}_i \in \mathbb{R}^3$ and $Z_i \in \mathbb{N}$ are the $i$-th atom's coordinate and atomic number, respectively,
$N$ is the number of atoms in the system,
$E$ is the total energy,
and $\vec{F}_i$ is the $i$-th atom's force.

Various representation styles have been used for NNPs, including simple rotation-invariant representations~\cite{NIPS2017_303ed4c6,zhang2018end,gasteiger_dimenet_2020},
a pair of rotation-invariant and -equivariant tensors~\cite{schutt2021equivariant},
a pair of node and edge embeddings~\cite{10.5555/3540261.3540781,gasteiger2022gemnetoc},
and $SO(3)$ tensors~\cite{liao2024equiformerv}.
A head transforms these representations into per-atom or per-system values to be predicted.
An NNP may have multiple heads to predict labels such as energy, forces, and stress~\cite{10.5555/3540261.3540781,gasteiger2022gemnetoc,liao2024equiformerv,barrosoluque2024openmaterials2024omat24}.
This architecture enables a simple fine-tuning method where pre-trained heads are reset while the corresponding encoder is reused at the beginning of fine-tuning.

\subsection{Large-scale material datasets}

Several DFT-labeled open datasets have been proposed for training NNP models.
One of the largest datasets currently available is Open Catalyst (OC), a family of datasets about catalysts~\cite{doi:10.1021/acscatal.0c04525,doi:10.1021/acscatal.2c05426,doi:10.1021/acscentsci.3c01629,barrosoluque2024openmaterials2024omat24}.
It consists of
OC20 (134M catalyst and adsorbate structures) and
OC22 (8.2M oxide catalyst and adsorbate structures),
along with some other large datasets developed by the group:
ODAC23 (36M metal-organic framework (MOF) and adsorbate structures)
and OMat24 (101M bulk structures).
Note that the numbers in the parentheses are for training with respect to their S2EF or S2EF-Total tasks.
Each OC sub-dataset provides various labels for one or more NNP tasks, including S2EF (structure to adsorption energy and forces), S2EF-Total (structure to total energy and forces), IS2RE (initial structure to relaxed energy), IS2RE-Total (initial structure to total energy at relaxed state).
Therefore, we choose the dataset and its training library as the foundation for our pretraining study.

Another example of a large dataset for materials science is PubChem, an open database containing 119M compounds' information (as of December 2024) such as SMILES (graph structural representations).
As a part of the database, PubChem3D provides conformers of PubChem compounds computed by the MMFF94s force fields, providing a wide variety of structural information on existing organic structures.
However, a challenge to utilizing PubChem3D for NNPs is that energy and force labels are not provided.
Therefore, techniques other than conventional supervised learning need to be introduced, such as denoising self-supervised learning, which is explained in \secref{subsection:denoising} in detail.

\subsection{Pre-Trained NNP models} \label{subsection:pretrained_models}
Pre-training of NNP models reduces dataset preparation and fine-tuning costs for applying them to specific materials to be investigated, just as in other machine learning tasks.
An example of transfer-learning of OC20-pre-trained models to the OC22 dataset and joint learning using both datasets is demonstrated in detail in the OC22 work; it implies that, even if a downstream task's material has different compositions (i.e., non-oxide and oxide catalysts) from those of the pre-training dataset, pre-training has a positive effect on the quality of fine-tuning.
Therefore, it is expected that pre-training a universal NNP model using every possible material dataset will result in an omnipotent foundation model for NNPs and, thus, computational materials science.

Examples of prior pre-training NNP model work and our work are shown in \tabref{table:related}.
These studies perform joint training by combining multiple datasets, such as organic and inorganic data, to feed abundant information to the models.
Many models adopt multi-head approaches to seamlessly combine multiple datasets, where an independent head (decoder) is used to predict the energy and force labels from each subset.
This is because each dataset adopts different DFT settings, including functionals, which determine the trade-off between the accuracy and computational cost of DFT calculations.
Several works, including PFP and DPA-2, use their own datasets by performing costly DFT calculations; other works, such as JMP and our work, utilize only open datasets to avoid the computational cost.
A unique aspect of our work among these pre-training approaches is that we use an unlabeled dataset, PubChem3D, as a subset to increase the amount of data.
The new method introduced to utilize the dataset is explained in \secref{subsection:loss}.
To the best of our knowledge, the \lamm{} joint dataset is one of the largest pre-training datasets for NNP models.

\begin{table}[tbhp]
  \centering
  \caption{
    Comparison of large-scale pre-trained NNP models,
    sorted in ascending lexicographical order for the right three columns.
  }
  \begin{tabular}{l|l r r r}
    \hline
    & Base model & \# params. & \# samples & Year \\
    \hline
    \hline
    OC20+OC22~\cite{doi:10.1021/acscatal.2c05426} & GemNet-OC~\cite{gasteiger2022gemnetoc} & 41.5M & 142M & 2022 \\
    MACE-MP-0~\cite{batatia2024foundationmodelatomisticmaterials} & MACE~\cite{NEURIPS2022_4a36c3c5} & 4.7M (middle) & 1.6M~\cite{deng2023chgnet} & 2023 \\
    DPA-2~\cite{zhang2024dpa2largeatomicmodel}    & DPA-2                                  & 7.68M &   4M & 2023 \\
    MatterSim~\cite{yang2024mattersimdeeplearningatomistic} & M3GNet~\cite{chen2022universal} & 4.5M & 17M & 2024 \\
    MatterSim~\cite{yang2024mattersimdeeplearningatomistic} & Graphormer~\cite{ying2021transformers} & 182M & 17M & 2024 \\
    PFP v7~\cite{pfpv7}                           & TeaNet~\cite{TAKAMOTO2022111280}       & Unknown &  59M & 2024 \\
    JMP-S~\cite{shoghi2024from}                   & GemNet-OC   &  30M & 120M & 2024 \\
    JMP-L~\cite{shoghi2024from}                   & GemNet-OC-L & 230M & 120M & 2024 \\
    \textbf{\lamms{}} (ours) & \painn{}~\cite{schutt2021equivariant} & 20.1M & 282M & 2025 \\
    \textbf{\lamml{}} (ours) & \eqvt{}~\cite{liao2024equiformerv}    & 33.8M & 282M & 2025 \\
    \hline
  \end{tabular}
  \label{table:related}
\end{table}

%% file: 3_proposal.tex
\begin{figure*}[tbhp]
  \centering
  \includegraphics[width=\linewidth]{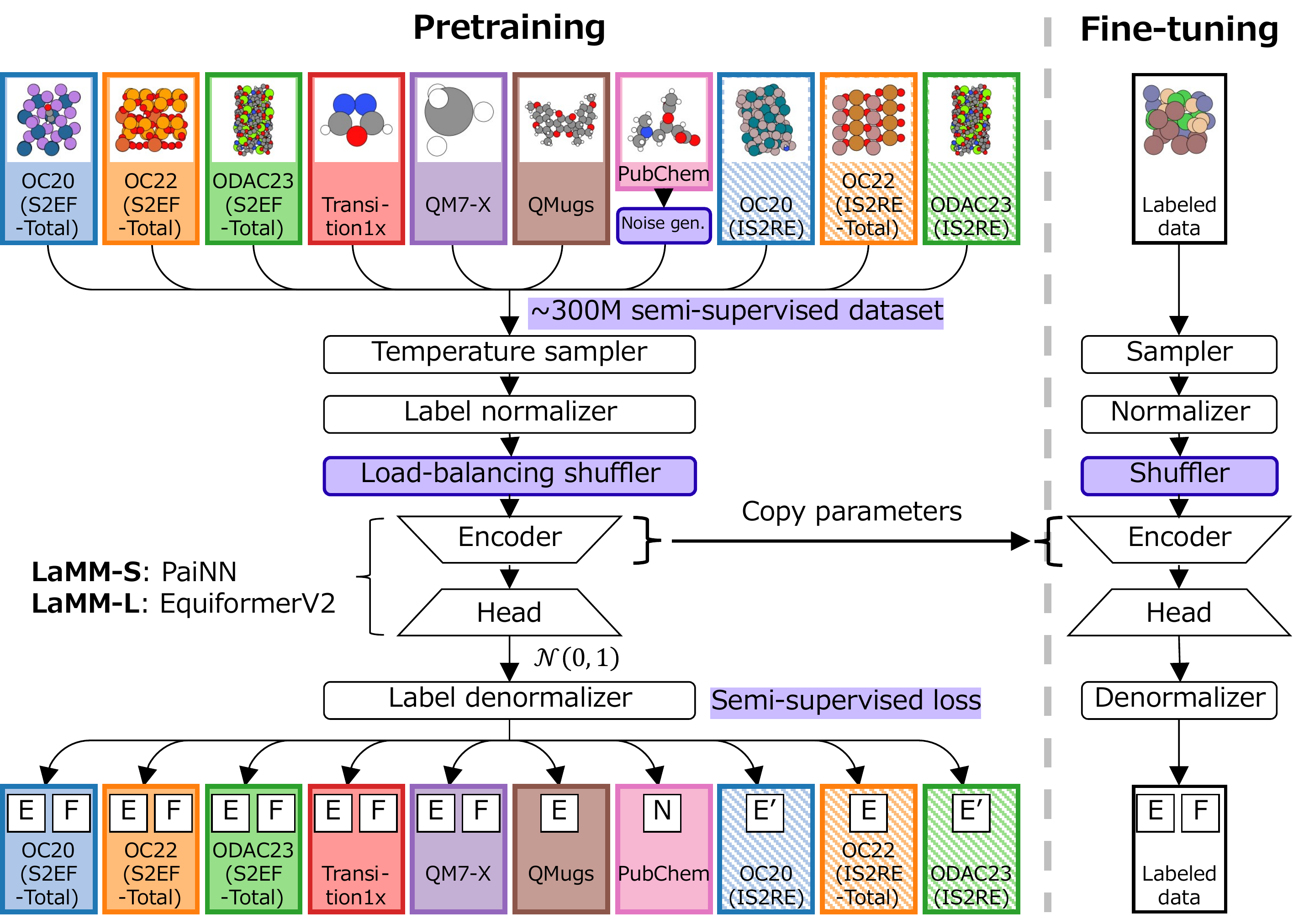}
  \caption{Overview of \lamm{}.}
  ``E'': total energy, ``E$'$'': adsorption energy after relaxation, ``F'': forces, ``N'': coordinate noises.
  The blue parts are our proposed methods explained in this paper.
  \label{figure:lamm}
\end{figure*}

\section{Proposal: Large Materials Model (\lamm{})} \label{section:proposal}

In this section, we explain the philosophy and methodology of our \textbf{Large Materials Model} (\textbf{\lamm{}}).
\lamm{}'s goal is to build a foundation model for various AI-driven material technologies,
such as screening and molecule dynamics simulations.

We adopt two different ML potential models, \painn{}~\cite{schutt2021equivariant} and \eqvt{}~\cite{liao2024equiformerv}, with different inference speeds, to construct our customized models, \textbf{\lamms{}} and \textbf{\lamml{}}, respectively.
This strategy supports both MD simulations, which require fast inference, and screening tasks, which require accurate single-point calculations.

The overview of \lamm{} is shown in \figref{figure:lamm}, which consists of the following components:

\begin{itemize}
\item A 300M-sample joint dataset consisting of multiple open datasets (\secref{subsection:dataset}).
\item \painn{} and \eqvt{} model heads and a loss function for semi-supervised joint training (\secref{subsection:loss}).
  They enable the training of a single model with any combination of system- and atom-labeled and unlabeled datasets.
  Especially in the material science field, the computational cost of labeling (DFT computation) is quite intensive.
  Therefore, this technique efficiently enlarges the pre-training dataset by an order of magnitude without additional computational cost.
\item A quality noise generation algorithm for self-supervised coordinate denoising tasks (\secref{subsection:denoising}).
  This algorithm improves convergence quality by applying unique labels to randomly deformed structure data.
\item A load-balancing algorithm for a dataset with large variance in the number of atoms on GPU supercomputers (\secref{subsection:loadbalancing}).
\end{itemize}

Their details are explained in the following sections.

\subsection{The 300M semi-supervised dataset} \label{subsection:dataset}
It is well reported that the inference accuracy of machine learning models is greatly affected by the dataset size used.
One of the most basic ways to improve the inference accuracy of ML models is to increase the dataset size, and similar reports have been made regarding ML potentials~\cite{doi:10.1021/acscatal.0c04525,doi:10.1021/acscatal.2c05426,doi:10.1021/acscentsci.3c01629}.
Specifically, a pre-training dataset should contain knowledge that correlates with possible downstream tasks.
Therefore, as part of our data collection policy, we decided to collect as diverse and large a number of input structures and labels as possible for pre-training \lamm{}.
Furthermore, we accept structures without labels rather than focusing on only labeled data, as studies exist on performing self-supervised training for machine learning potentials~\cite{feng2023fractional}.

The overview of our pre-training dataset is shown in \tabref{table:datasets}.
Our dataset comprises 10 open datasets of various materials, including catalysts and small molecules, with a total of around 300M data samples.
A difference between our dataset and other pre-training datasets is that around 1/3 of it is occupied by PubChem3D~\cite{bolton2011pubchem3d}, an unlabeled molecule dataset,
whose molecule structures are optimized by the MMFF94s force field.
To utilize this large dataset source, we propose a loss function and an effective denoising labeling method to perform semi-supervised pre-training.

We follow the methodology of JMP~\cite{shoghi2024from}, i.e., apply temperature sampling ($T=2$) to our pre-training dataset to mitigate the difference in the number of samples between its subsets.
We fix the number of samples of the largest subset, OC20 (S2EF-Total), and repeat other samples in each epoch to satisfy the proportion of the temperature sampling.

We use the training and validation data splits of the OC20, OC22, and ODAC23 subsets as is; for the other subsets, we split them into 99\% and 1\% for training and validation, respectively.
We remove data samples with more than 300 atoms from ODAC23 because they usually cause GPU out-of-memory errors.

\begin{sidewaystable}
  \small
  \centering
  \caption{
    Overview of our pre-training dataset.
    ``E'': Energy, ``F'': Forces, ``N'': Pseudo-force labels.
    ``S2EF-T'': S2EF-Total.
    ``IS2RE-T'': IS2RE-Total.
    ``TS'': Temperature sampling with $T=2$.
    *: Data samples of more than 300 atoms are excluded from ODAC23, as explained in \secref{subsection:dataset}.
  }
  \begin{tabular}{l|c c c c c r r r r}
    \hline
    & \multirow{2}{*}{Domain} & \multirow{2}{*}{Task} & \multirow{2}{*}{DFT functional} & \multirow{2}{*}{Periodic} & \multirow{2}{*}{Labels} & \# avg. & \multirow{2}{*}{Elements} & \multirow{2}{*}{\# samples} & \# samples \\
    &&&&&& atoms &&& w/ TS \\
    \hline
    \hline

    OC20 (S2EF-T) ~\cite{doi:10.1021/acscatal.0c04525}     & Catalyst       & S2EF-T & RPBE       & \checkmark{} & E, F &  73 & 56 & 134M & 134M \\
    OC22 (S2EF-T) ~\cite{doi:10.1021/acscatal.2c05426}     & Oxide catalyst & S2EF-T & PBE+U      & \checkmark{} & E, F &  80 & 57 & 8.2M &  33M \\
    ODAC23* (S2EF-T) ~\cite{doi:10.1021/acscentsci.3c01629} & MOF catalyst   & S2EF-T & PBE+D3     & \checkmark{} & E, F & 160 & 70 &  29M &  63M \\
    Transition1x~\cite{schreiner2022transition1x} & Molecule & S2EF-T & $\omega$B97x/6–31 G(d)     & \crossmark{} & E, F &  14 &  4 & 9.5M &  36M \\
    QM7-X~\cite{hoja2021qm7}                      & Molecule & S2EF-T & PBE0+MBD                   & \crossmark{} & E, F &  17 &  6 & 4.2M &  24M \\
    QMugs~\cite{isert2022qmugs}                   & Molecule & S2EF-T & $\omega$B97X-D/def2-SVP    & \crossmark{} & E    &  55 & 10 & 2.0M &  16M \\
    PubChem3D~\cite{bolton2011pubchem3d}            & Molecule & Denoising  & N/A                     & \crossmark{} & N    &  46 & 11 &  94M & 112M \\
    OC20 (IS2RE)   & Catalyst       & IS2RE   & RPBE   & \checkmark{} & E &  78 & 56 & 460k & 7.9M \\
    OC22 (IS2RE-T) & Oxide catalyst & IS2RE-T & PBE+U  & \checkmark{} & E &  84 & 57 &  46k & 2.5M \\
    ODAC23 (IS2RE) & MOF catalyst   & IS2RE   & PBE+D3 & \checkmark{} & E & 162 & 70 & 132k & 4.2M \\
    \hline
    Total & & & & & & & 79 & 282M & 432M \\
    \hline
  \end{tabular}
  \label{table:datasets}
\end{sidewaystable}

\begin{figure*}[tbhp]
  \centering
  \includegraphics[width=1.5\linewidth,angle=90]{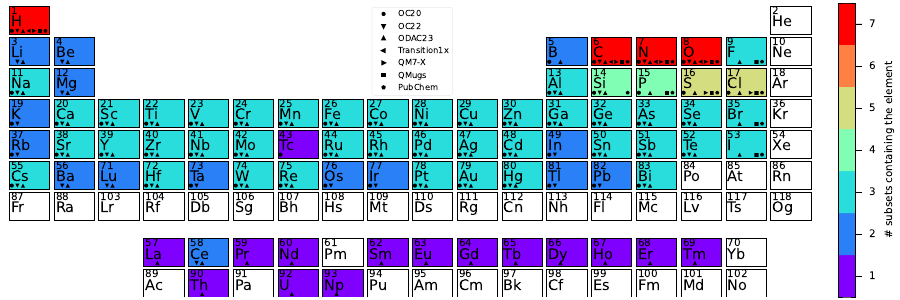}
  \caption{
    Heatmap of the number of datasets containing each element.
    Markers represent which subsets contain the corresponding element.
  }
  \label{figure:periodic_table}
\end{figure*}

\figref{figure:periodic_table} shows the elements that appear in the subsets.
H, C, N, and O are included in all the subsets, and most of the other elements, except for noble gases and elements beyond the seventh period, are covered by OC20, OC22, and ODAC23.

The dataset contains energy and force labels for four different types of tasks:
S2EF-Total (structure to total energy and forces)~\cite{doi:10.1021/acscatal.2c05426},
IS2RE (input structure to the relaxed adsorption energy)~\cite{doi:10.1021/acscatal.0c04525},
IS2RE-Total (input structure to relaxed total energy)~\cite{doi:10.1021/acscatal.2c05426},
and coordinate denoising.
Using this diverse knowledge to train models makes the resulting pre-trained models easily fit similar tasks, as demonstrated in \secref{section:evaluation}.
Note that the IS2RE and IS2RE-Total datasets are omitted from \figref{figure:periodic_table}, \figref{figure:datasetnatoms}, and \figref{figure:featuremap},
as their structures are similar to those of the corresponding S2EF-Total datasets.

\subsection{Universal loss and head design} \label{subsection:loss}
In this section, we propose a loss function for semi-supervised joint training of machine learning potentials.
This function accepts any combination of datasets containing
system-wide energy and per-atom forces (e.g., OC20),
energy only (e.g., QMugs),
and per-atom forces (e.g., PubChem3D, where forces are randomly generated as explained in \secref{subsection:denoising}),
to train a single model.

Let $x_i$ be the input data (atom coordinates and elements) of the $i$-th data sample in a mini-batch of size $B$,
$N_i$ be the number of atoms in the sample,
$\hat{E}_i$ be its energy ($\hat{E}_i = 0$ if unavailable),
$\{{\hat{F}_i^{(j)}}\}_{j=1}^{N_i}$ be its forces ($\hat{F}_i^{(j)} = \vec{0}$ for any $j$ if unavailable), and
$d_i$ be its integer dataset index,
$m_E^{(i)}$ and $m_F^{(i)}$ be 0 or 1 masks indicating whether the sample contains energy and force labels, respectively.

\lamm{} predicts its energy as if the sample belonged to each subset,
$E_i = f_E(x_i|\theta) \in \mathbb{R}^{|D|}$, where $|D|$ is the number of subsets ($|D|=10$ in pre-training) and $\theta$ is the current parameters.
Similarly, it predicts forces $\{F_i^{(j)}\}_{j=1}^{N_i} = f_F(x_i|\theta)$, where $F_i^{(j)} \in \mathbb{R}^{3 \times |D|}$ for any $j$.
Even though it is unnecessary to compute energy and forces for all the subsets because some of the target labels are missing, this approach batches the computation of the final layer of energy and force heads for each dataset, especially for \lamm{}, whose output layer is a linear layer.
We set the number of output channels in the final fully connected layers of \lamms{} and \lamml{} to a multiple, allowing for the inference of energy and force for each subset.

The loss $\mathcal{L}$ of the mini-batch $\{(x_i, \hat{E}_i, \{{\hat{F}_i^{(j)}}\}_{j=1}^{N_i}, d_i, m_E^{(i)}, m_F^{(i)})\}_{i=1}^{B}$ is defined as follows:
\begin{eqnarray}
  \mathcal{L} &=& \frac{\lambda_E}{\sum_{i=1}^{B} m_E^{(i)}} \sum_{i=1}^{B} m_E^{(i)} | \hat{E}_i - (E_i)_{d_i} | \nonumber \\
  &+& \frac{\lambda_F}{\sum_{i=1}^{B} m_F^{(i)}} \sum_{i=1}^{B} \frac{m_F^{(i)}}{N_i} \sum_{j=1}^{N_i} || \hat{F}_i^{(j)} - \{F_i^{(j)}\}_{d_i} ||_2
\end{eqnarray}
where $\lambda_E$ and $\lambda_F$ are coefficients of energy and force losses, respectively.
We use MAE loss for energy and system size-balanced L2 MAE losses~\cite{shoghi2024from} for forces in this loss function.
We also apply the linear reference energy technique introduced in the OC22 work~\cite{doi:10.1021/acscatal.2c05426} and normalization to both energy and forces, as mentioned in the JMP work~\cite{shoghi2024from}.

\subsection{Efficient denoising labeling} \label{subsection:denoising}
Prior work proposes training machine learning potentials as a self-supervised ``denoising'' task, where random noise is added to atom coordinates and the noise is set as the corresponding label~\cite{feng2023fractional}.
The motivation for such a denoising method is two-fold: 1) Training can be done without energy and force labels, which require much computational cost, and 2) the objective of denoising strongly correlates to the force field~\cite{feng2023fractional}.
In this work, we introduce a simple technique to improve the convergence speed and accuracy of such a denoising method.

\begin{figure}[tbhp]
  \centering
  \subfigure[Existing]{
    \includegraphics[width=0.4\linewidth]{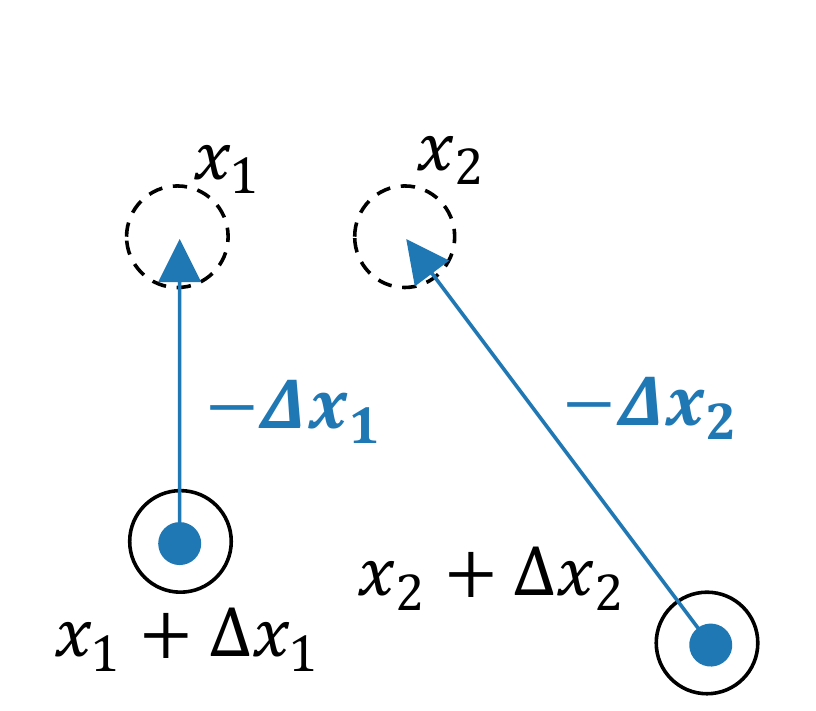}
    \label{figure:denoising_existing}
  }
  \subfigure[Proposal]{
    \includegraphics[width=0.5\linewidth]{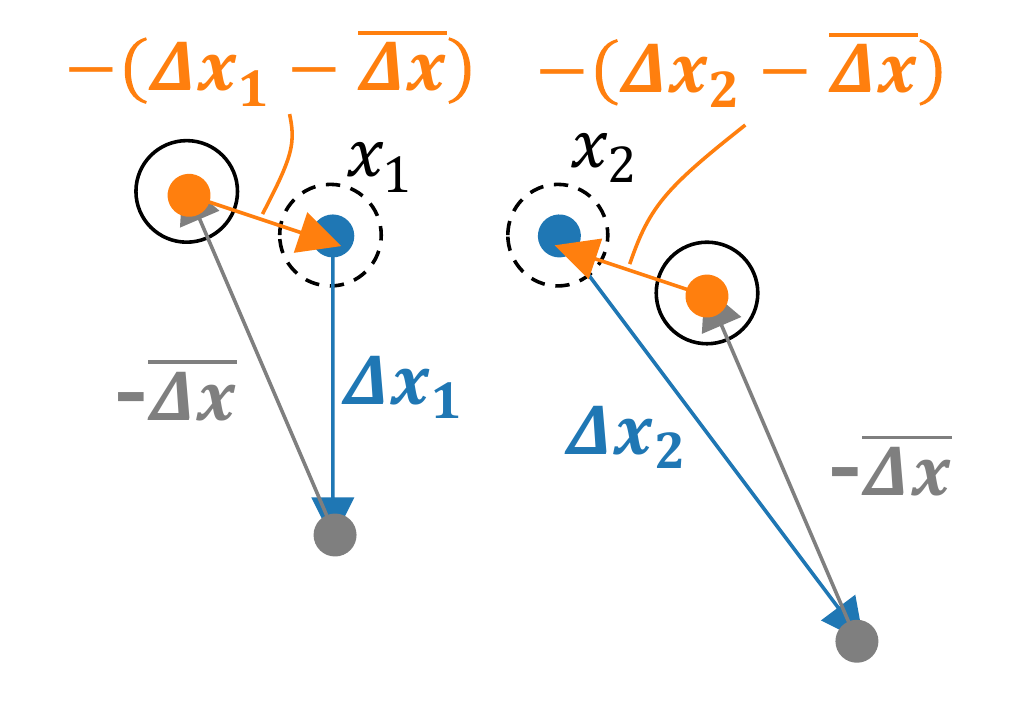}
    \label{figure:denoising_proposed}
  }
  \caption{Existing and  proposed denoising labeling algorithm.
    \textbf{(a)} Input coordinate $x_i + \Delta x_i$ and labels $-\Delta x_i$ of the existing method.
    \textbf{(b)} Input coordinate $x_i + \Delta x_i - \overline{\Delta x}$ and labels $-(\Delta x_i - \overline{\Delta x})$ of the proposed method.
  }
  \label{figure:denoising}
\end{figure}

\figref{figure:denoising} shows the overview of existing denoising and the proposed denoising tasks.
In \figref{figure:denoising_existing},
a 3D Gaussian noise vector $\Delta x_i \in \mathbb{R}^3$ is generated and added to atom $i$'s coordinate $x_i \in \mathbb{R}^3$, and a model is trained to predict the noise $\Delta x_i$ from the new input coordinate $x_i + \Delta x_i$.
For simplicity, in this work, we regard $-\Delta x_i$ as force labels so that the denoising task is naturally combined with other labeled datasets, as explained in \secref{subsection:loss}.

However, in this method, labels are not unique to pairs of the ground-truth atom coordinates and input coordinates.
For instance, in \figref{figure:denoising_existing}, the direction of both labels is upward, and there is arbitrariness in their translational component $- \overline{\Delta x} = - N^{-1} \sum_{i=1}^{N} \Delta x_i$ where $N$ is the number of atoms.
Since many potential models take atoms' relative coordinates as input, this method implicitly adds such arbitrary offsets to labels, which leads to worse convergence.

In this work, we propose a denoising task that defines $\Delta x_i - \overline{\Delta x}$ as the noise to be added to solve this problem (\figref{figure:denoising_proposed}).
In this method, unique labels corresponding to the generated noise vectors are assigned to atom coordinates, which results in better convergence.

\subsection{Efficient load balancing for GPUs} \label{subsection:loadbalancing}

Load balancing is a crucial performance factor for training ML models with a dataset of diverse data sample sizes.
For example, the forward and backward computation time of most machine learning potentials is strongly correlated with the number of atoms in an input system. This may lead to load imbalance in multi-GPU training with samples whose number of atoms varies by orders of magnitude.
This problem becomes critical, especially when the GPU batch size is restricted to a few samples due to the complexity of the model, such as \eqvt{}.

\begin{figure}[tbhp]
  \centering
  \includegraphics[width=\linewidth]{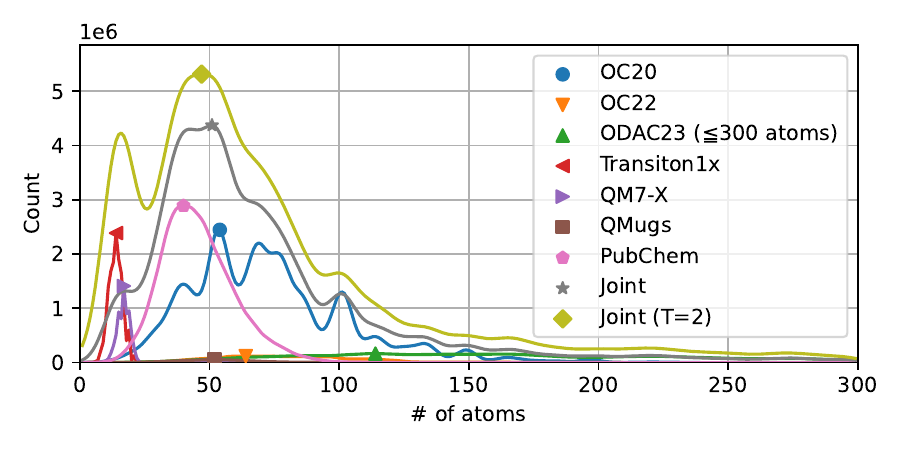}
  \caption{
    Histogram of the per-sample number of atoms of our pre-training dataset.
    Each histogram is kernel density estimated using Gaussian kernels with $10^5$ randomly sampled structures.
    Each point represents the location where the corresponding histogram is maximized.
  }
  \label{figure:datasetnatoms}
\end{figure}

As an example of such a diverse dataset, the distribution of the number of atoms for each subset of our pre-training dataset is shown in \figref{figure:datasetnatoms}.
As the markers show, the peaks of the distributions range from around 10 atoms to more than 100 atoms.
The temperature sampling technique worsens the situation by increasing the virtual size of small subsets such as Transition1x and QM7-X, whose number of atoms is much smaller than the average number of atoms in the original joint datasets.

We found three performance issues when training \lamm{} with this problem:

\begin{itemize}
\item A small per-GPU batch size increases the variance of the per-GPU workload ($\sim$ the total number of atoms per GPU).
  This situation makes some GPUs idle until all GPUs complete forward and backward computation, before beginning all-reduce communication of gradients, which degrades the effective computational performance.
  Simply fixing the ratio of samples to be sampled from each subset cannot solve the issue because, as seen in \figref{figure:datasetnatoms}, each subset may contain data samples with a wide range in the number of atoms, from $O(10^1)$ to $O(10^2)$.
  Also, it is worth mentioning that \fairchem{} implements heuristics to greedily assign data samples of a mini-batch to GPUs one by one to minimize the difference in the workload (the number of atoms or graph edges) among GPUs~\cite{ocp267}.
  However, this approach cannot solve the problem if the GPU batch size is limited to 1 or a few samples for the cases of large models such as \eqvt{}.

\item Having multiple large samples occasionally in a GPU batch may lead to an out-of-memory error.
  Whether the error occurs depends on the result of dataset shuffling. Hence, we need to introduce a mechanism to avoid generating such an imbalanced batch for all GPUs.

\item If larger data samples with respect to memory consumption are fed to the network than in the previous step, the machine learning framework (PyTorch in this work) may invoke costly GPU memory reallocation.
  As explained in \secref{subsection:load_balancing_eval}, this overhead is one of the crucial performance issues in large-scale pre-training.
\end{itemize}

\begin{figure}[tbhp]
  \centering
  \includegraphics[width=\linewidth]{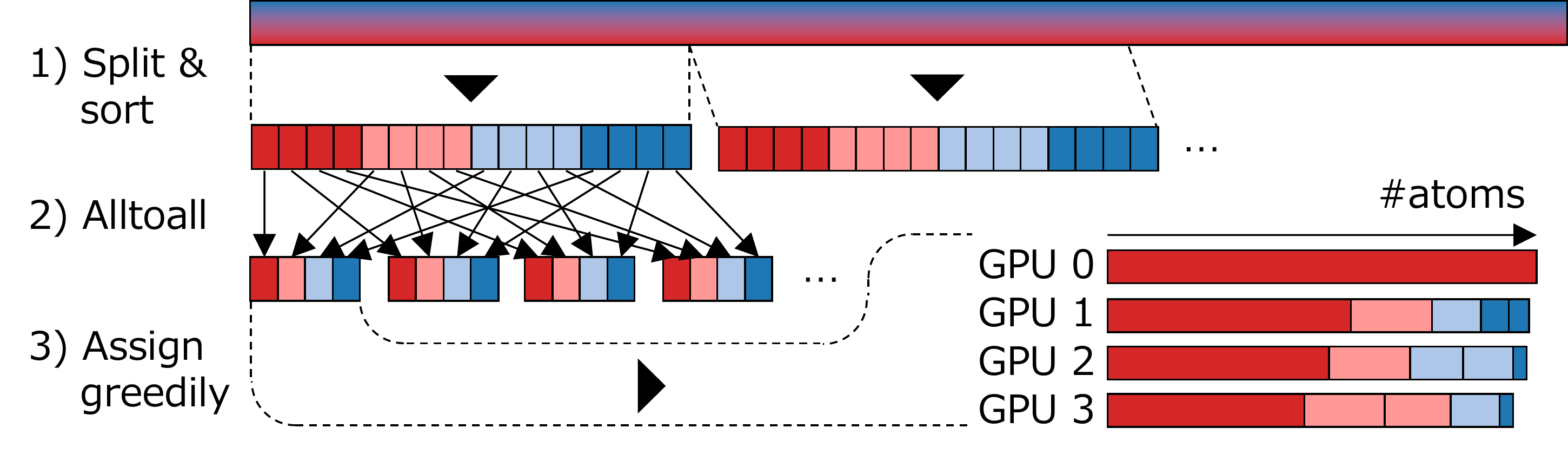}
  \caption{
    Our proposed load balancing algorithm.
    Red/blue samples contain a large/small number of atoms, respectively.
    On bottom-right, a mini-batch of 16 samples are assigned to 4 GPUs.
  }
  \label{figure:loadbalancing}
\end{figure}

To resolve this load-imbalance problem, we propose a rearrangement method based on the number of atoms before training (\figref{figure:loadbalancing}).

The method comprises the following three steps:

\begin{enumerate}
\item The entire dataset is split into a given number of subsets, and each subset is sorted in descending order according to the number of atoms.

\item Each sorted subset is split into chunks whose number of samples is equal to the number of GPUs, and an all-to-all rearrangement is applied to the chunks to construct mini-batches.

\item A greedy approach is applied to each mini-batch to assign samples to GPUs.
\end{enumerate}

The motivation for splitting in step 1 is to mitigate the temporal locality of each subset during training; for example, as shown in \figref{figure:datasetnatoms}, ODAC23 is composed of relatively large data samples, and hence many of the samples are used at the beginning of training without the splitting.
Similarly, step 2 is performed based on the same motivation instead of simple sorting.

Another trick in step 1 is that the descending sort guarantees that the total number of atoms in each mini-batch is smaller than that of the previous mini-batch, which reduces the memory rearrangement overhead.

%% file: 4_evaluation.tex
\section{Evaluation} \label{section:evaluation}

In this section, we investigate the computational performance of our new methods, the result of our pre-training and its analysis, and the fine-tuning performance for several datasets that are not included in our pre-training dataset.

We use the ABCI 2.0 supercomputer~\cite{takizawa2021abci} for pre-training and performance measurement.
We use its (V) computing nodes, each of which is equipped with four NVIDIA V100 GPUs with 16 GiB HBM2 memory, two Intel Xeon Gold 6148 CPUs, and 32 GiB DDR4 memory, and is connected with EDR InfiniBand.
The pre-training dataset is stored in its Lustre network file system and loaded to CPU memory with 20 and 10 workers per process for \lamms{} and \lamml{}, respectively.
We also use a single-node machine equipped with two NVIDIA Tesla P100 GPUs for the fine-tuning experiment.

We implement our proposed methods of \secref{section:proposal} in the \fairchem{} (formerly OCP) library~\cite{doi:10.1021/acscatal.0c04525} (commit \texttt{9a1f866}), combined with PyTorch 2.4.0 and CUDA 12.4.

\subsection{Effect of our load-balancing algorithm} \label{subsection:load_balancing_eval}

\newcommand{\loadbalancinglammlslowdown}[0]{1.64x}

\begin{figure}[tbhp]
  \centering
  \subfigure[\lamms{} (step)]{
    \includegraphics[width=0.45\linewidth]{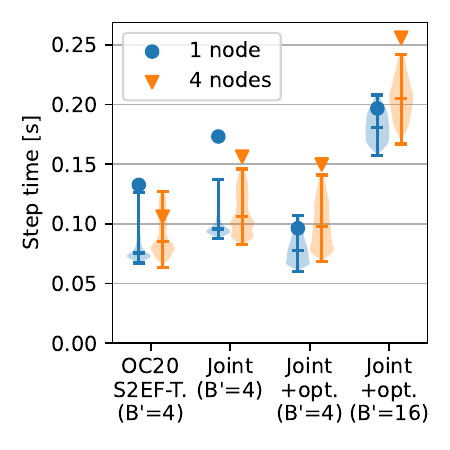}
    \label{figure:load_balancing_eval_lamm_s}
  }
  \subfigure[\lamml{} (step)]{
    \includegraphics[width=0.45\linewidth]{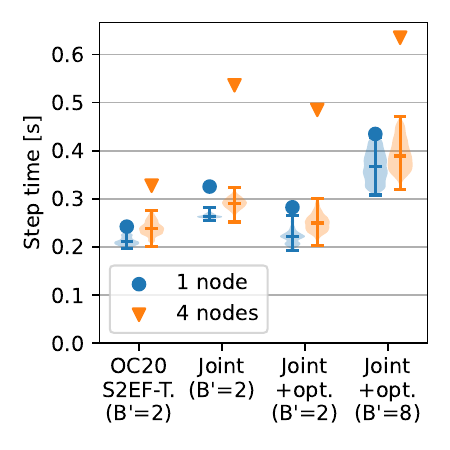}
    \label{figure:load_balancing_eval_lamm_l}
  }
  \subfigure[\lamms{} (forward+backward, 4 nodes)]{
    \includegraphics[width=0.99\linewidth]{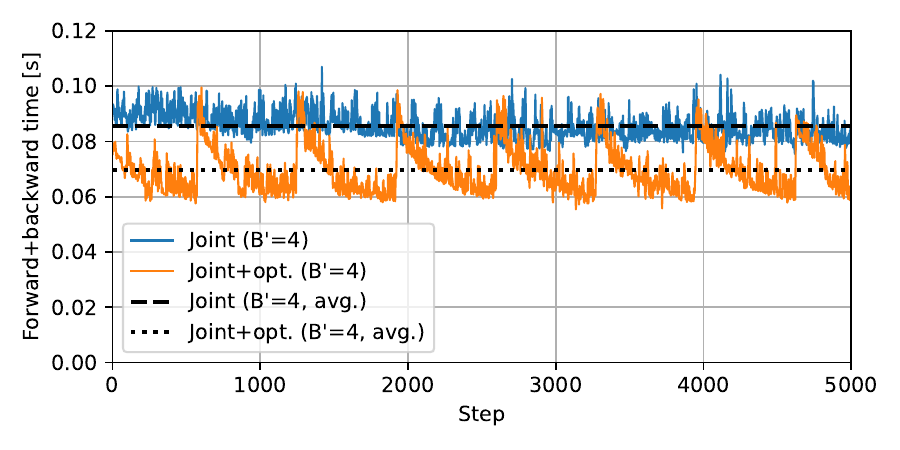}
    \label{figure:load_balancing_eval_lamm_s_forward_backward}
  }
  \caption{
    Training step time with and without our load-balancing method on ABCI (V).
    ``S2EF-T.'': S2EF-Total, ``Joint'': the joint dataset with temperature sampling of \tabref{table:datasets}, `$B'$: GPU batch size, ``opt.'': the proposed method.
    Each dataset is split into 10,000 subsets in the first step of the proposed method.
    In \figref{figure:load_balancing_eval_lamm_s} and \figref{figure:load_balancing_eval_lamm_l}, data points smaller than the 95th percentile of the entire data are shown for the violin plot, as the distribution exhibits a long tail.
    Each violin's ticks represent its minimum, median, and maximum values, respectively, and the marker represents its average, including outliers.
    In \figref{figure:load_balancing_eval_lamm_s_forward_backward}, the first 100 steps are removed and 5-point moving averaging is applied for visibility.
  }
  \label{figure:load_balancing_eval}
\end{figure}

\figref{figure:load_balancing_eval_lamm_s} and \figref{figure:load_balancing_eval_lamm_l} show the training step time distribution of \lamms{} and \lamml{} on the ABCI supercomputer, respectively.
We evaluate two different datasets, OC20 (S2EF-Total) and our joint dataset with temperature sampling ($T=2$), to investigate how the variance in the number of atoms affects the computational performance.
The figures show that, especially with \lamml{}, the variance slows down the training.
The average step time of \lamml{} on 4 nodes is \loadbalancinglammlslowdown{} slower when changing the dataset from OC20 to the joint dataset, even though the average number of atoms is not increased as shown in \tabref{table:datasets}.
The reason for the slowdown includes load imbalances and memory allocation as mentioned in \secref{subsection:loadbalancing}.

Our load-balancing algorithm, denoted as ``opt.'', improves the throughput by at most \loadbalancinglammsspeedup{} with \lamms{} on one node (4 GPUs).
\figref{figure:load_balancing_eval_lamm_s_forward_backward} explains how the method improves the throughput.
Without our method, the sum of forward and backward time, including all-reduce communication, is limited by the slowest GPU of each step, and hence the time is almost constant throughout training.
On the other hand, our method minimizes the load imbalance among GPUs, so the step time gradually decreases as the total number of atoms within each split subset decreases.
This effect leads to acceleration in the average step time, as shown in \figref{figure:load_balancing_eval_lamm_s} and \figref{figure:load_balancing_eval_lamm_l}.

Another benefit of the method is that it avoids occasional GPU out-of-memory errors by removing many too-large samples in GPU batches, so users can safely increase the mini-batch size to increase the throughput;
we achieve \loadbalancinglammsthroughput{} and \loadbalancinglammlthroughput{} throughput increase on 4 nodes for \lamms{} and \lamml{}, respectively.

\subsection{Effect of our denoising labeling algorithm}

\newcommand{\denoisingspeedupasisepoch}[0]{1.36 epoch}
\newcommand{\denoisingspeedupoptepoch}[0]{0.68 epoch}
\newcommand{\denoisingspeedup}[0]{2.00x}
\newcommand{\denoisingspeedupmae}[0]{0.03~\AA}

\begin{figure}[tbhp]
  \centering
  \includegraphics[width=\linewidth]{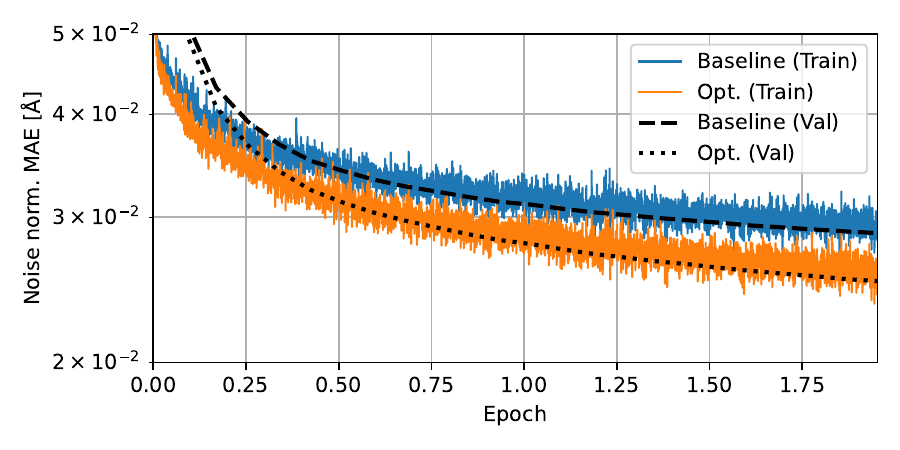}
  \caption{
    Learning curves for self-supervised training on the PubChem3D Compounds dataset.
    ``Baseline'': The baseline method shown in \figref{figure:denoising_existing}.
    ``Opt.'': The proposed method shown in \figref{figure:denoising_proposed}.
  }
  \label{figure:denoisingeval}
\end{figure}

\figref{figure:denoisingeval} shows the learning curves of denoising SSL on the PubChem3D dataset with and without our proposed method.
Our method reduces validation errors more than the baseline with the same number of steps;
for instance, we achieve \denoisingspeedup{} acceleration in reaching a validation error of \denoisingspeedupmae{} compared to the baseline (\denoisingspeedupasisepoch{} vs. \denoisingspeedupoptepoch{}).
This result demonstrates that our simple method effectively improves the quality of SSL with almost no additional computational cost.

\subsection{Pre-Training \lamm{}}
We performed large-scale pre-training of \lamms{} and \lamml{} with 64 V100 GPUs on 16 ABCI (V) computing nodes.
Since our main goal is to obtain a lightweight pre-trained model for multiple purposes, such as screening and molecule dynamics, we trained \lamms{} for two epochs and experimentally trained \lamml{} for 0.16 epochs.

\begin{figure*}[tbhp]
  \centering
  \includegraphics[width=\linewidth]{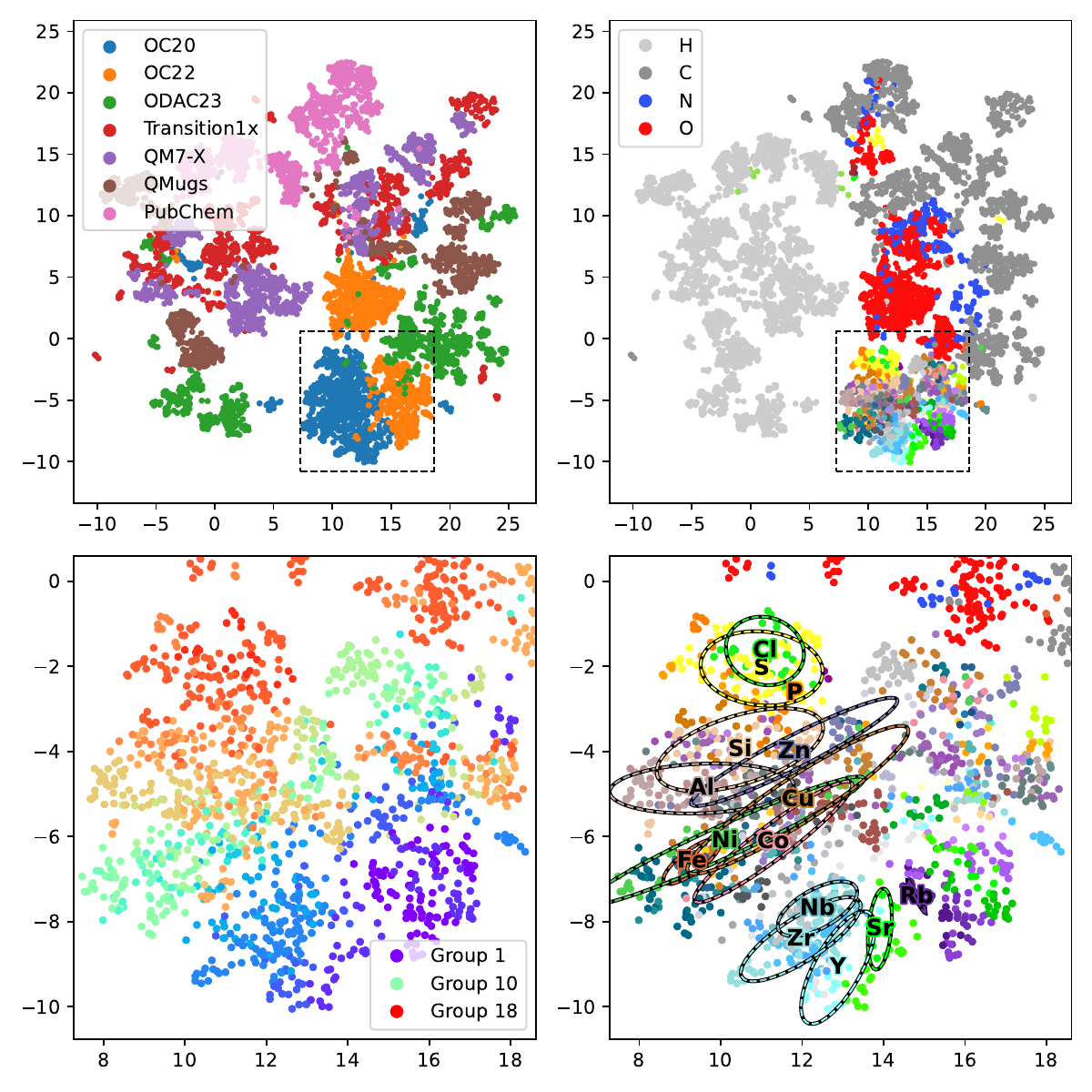}
  \caption{
    UMAP~\cite{mcinnes2018umap-software} node-level feature map of the pre-training dataset.
    \painn{}'s rotation-invariant representations with a length of 512 dimensions are used.
    1000 atoms are randomly sampled from each subset.
    \textbf{Top left}: colored by subset.
    \textbf{Top right}: colored by element with colors defined by the Jmol library~\cite{jmol}.
    \textbf{Bottom left}: elements other than H, C, N, and O (dashed areas in the top figures),
    colored by group.
    \textbf{Bottom right}: colored by element. Dashed ellipses and element names represent the $1.5 \sigma$ confidence regions using Pearson correlation coefficients and the average positions for manually chosen OC20 elements, respectively.
    P's ellipse is omitted for visibility.
  }
  \label{figure:featuremap}
\end{figure*}

\figref{figure:featuremap} visualizes atom-level representations of the joint dataset computed by \lamms{}.
We use PaiNN's rotation-invariant representations $s_i$ for the visualization, which are computed from relative coordinates and elements of nearby atoms.
The figure shows that the combination of the joint dataset and the model represents diverse interatomic structures, especially for organic materials' elements (H, C, N, and O).
For instance, hydrogen atoms are included in OC20, OC22, and ODAC23 as adsorbates, ODAC23 as part of MOFs, and other subsets as part of molecules.
The figure shows that the model distinguishes these interatomic structures by using its encoder without labels such as energy and forces.

Another finding is that the model learns the similarity of element properties.
In the figure, we highlight five manually chosen elements from the
third period
({\textcolor[rgb]{0.674,0.586,0.586}{\textbf{Al}}},
{\textcolor[rgb]{0.847,0.706,0.564}{\textbf{Si}}},
{\textcolor[rgb]{0.9,0.452,0.0}{\textbf{P}}},
{\textcolor[rgb]{0.9,0.9,0.169}{\textbf{S}}},
{\textcolor[rgb]{0.11,0.847,0.11}{\textbf{Cl}}}),
fourth period
({\textcolor[rgb]{0.79,0.36,0.18}{\textbf{Fe}}},
{\textcolor[rgb]{0.847,0.508,0.564}{\textbf{Co}}},
{\textcolor[rgb]{0.283,0.734,0.283}{\textbf{Ni}}},
{\textcolor[rgb]{0.706,0.452,0.18}{\textbf{Cu}}},
{\textcolor[rgb]{0.441,0.452,0.621}{\textbf{Zn}}}),
and fifth period
({\textcolor[rgb]{0.395,0.162,0.621}{\textbf{Rb}}},
{\textcolor[rgb]{0.0,0.9,0.0}{\textbf{Sr}}},
{\textcolor[rgb]{0.522,0.9,0.9}{\textbf{Y}}},
{\textcolor[rgb]{0.522,0.79,0.79}{\textbf{Zr}}},
{\textcolor[rgb]{0.406,0.685,0.709}{\textbf{Nb}}}),
respectively.
As these distributions show, the model encodes elements in nearby positions in the periodic table into similar representations.
This observation resembles the training results of other existing work~\cite{shoghi2024from,zhang2024dpa2largeatomicmodel}, which proves that our model learns universal knowledge about materials, but with a broader scope of materials.

\subsection{Fine-tuning \lamm{}}
In this section, we evaluate the fine-tuning performance of our pre-trained models for an unseen dataset (i.e., structures and a DFT functional not present in the pre-training dataset).
The primary goal of \lamm{} is not to provide a general-purpose machine learning potential for any input without additional cost, but rather to provide a general-purpose pre-trained model that can be fine-tuned for any input.
As shown in \tabref{table:datasets}, users demand that various DFT configurations, such as functionals, be applied to given inputs.
Therefore, the fine-tuning approach results in a relatively smaller cost, including both pre-training and fine-tuning, than constructing a massive pre-trained model that can predict properties for any DFT functionals.

\begin{figure}[tbhp]
  \centering
  \subfigure[Energy]{\includegraphics[width=0.4\linewidth]{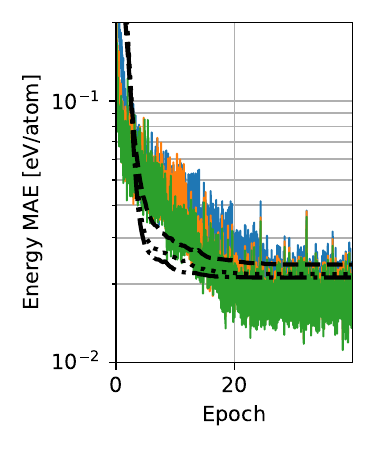}} 
  \subfigure[Forces]{\includegraphics[width=0.55\linewidth]{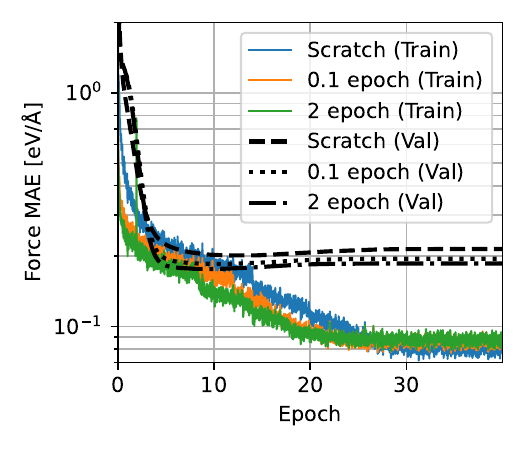}}
  \caption{
    Learning curves of \lamms{} for HME21 fine-tuning.
    31-points moving averaging is applied to the training curves for visibility.
  }
  \label{figure:finetuning}
\end{figure}

\figref{figure:finetuning} shows energy and forces learning curves on fine-tuning of \lamms{} for the HME21 dataset~\cite{takamoto2022towards}.
HME21 is an artificial DFT dataset containing multiple elements in disordered coordinates with PBE total energy and force labels per sample.
We compare the fine-tuning performance of three different \painn{} checkpoints of the \lamms{} pre-training at 0, 0.1, and 2 epochs.

The figure shows that the best validation errors of energy and forces improve as more pre-training is performed on the model.
Another observation is that the accuracy improvement between the 0.1 and 2 epoch versions is less significant than that between the 0 and 0.1 epoch versions, which implies that the 0.1 epoch model already has sufficient knowledge to improve the convergence of fine-tuning.
This result justifies our data strategy of collecting diverse data as much as possible, which is effective for fine-tuning the model for unseen datasets.

\tabref{table:finetuning} shows the fine-tuning performance of \lamm{} and other models for the HME21 dataset.
Note that the \painn{} model, which is the base model of \lamms{}, is a 6-layer model with an increased number of hidden channels defined in \fairchem{}, while the original model has three layers.
The table shows that both \lamms{} and \lamml{} improve the best accuracy for both datasets.
In addition, they reduce the number of fine-tuning steps to reach the given thresholds by half or less for most cases.
These results prove that the pre-training models successfully acquired universal knowledge for downstream tasks.

\begin{table*}[tbhp]
  \centering
  \caption{
    Fine-tuning performance of \lamms{} (\painn{}, 6-layers) and \lamml{} (\eqvt{}).
    Bold font indicates the best result within the group of lines containing \lamm{} models.
    ``Speedup'': Relative speedup compared with the corresponding base model to achieve the threshold in parentheses.
  }
  \begin{tabular}{l|r r}
    \hline
    & \multicolumn{2}{c}{HME21} \\
    & Energy MAE              & Forces MAE             \\
    & [meV/atom] $\downarrow$ & [meV/\AA] $\downarrow$ \\
    \hline
    \hline
    \painn{} (3 layers)~\cite{takamoto2022towards}  & 22.9 & 237 \\
    TeaNet~\cite{takamoto2022towards}            & 19.6 & 174 \\
    \hline
    \painn{}             & 23.7 & 201 \\
    \lamms{} (0.1 epoch) & 21.8 & 185 \\
    \lamms{} (2 epoch)   & \bestperf{21.1} & \bestperf{176} \\
    \multirow{2}{*}{Speedup by \lamms{}} & 2.51x & 1.89x \\
    & (25 meV/atom) & (210 meV/\AA) \\
    \hline
    \eqvt{} (31M)        & 24.4 & 136 \\
    \lamml{} (0.16 epoch)& \bestperf{20.2} & \bestperf{115} \\
    \multirow{2}{*}{Speedup by \lamml{}}  & 2.50x & 2.61x \\
    & (25 meV/atom) & (150 meV/\AA) \\
    \hline
  \end{tabular}

  \label{table:finetuning}
\end{table*}

%% file: 5_conclusion.tex
\section{Conclusion}
In this work, we introduced \lamm{}, an omnipotent foundation NNP model trained with 300M data samples.
We also demonstrated that our new training methodology enables NNP training with any material dataset, even if some labels are missing.
This method increases the amount of available data for training, which improves the quality of NNPs.
Although this work demonstrated our semi-supervised learning approach only for pre-training, it can also be applied to fine-tuning and other downstream tasks to reduce DFT labeling costs.
Therefore, we believe that our proposed method will lead to a broader and more accurate application of NNPs for computational materials science.